\documentclass[runningheads]{llncs}

\usepackage{eccv}

\usepackage{eccvabbrv}

\usepackage{graphicx}
\usepackage{booktabs}
\usepackage{xcolor}
\usepackage{colortbl}

\usepackage[accsupp]{axessibility}  

\usepackage{hyperref}

\usepackage{orcidlink}
\usepackage{adjustbox}

\begin{document}

\title{Monte Carlo Energy Aggregation for Mobile 3D Gaussian Splatting  } 

\titlerunning{Flux-GS: Monte Carlo Energy Aggregation}

\author{Xiaobiao Du\inst{1, 2} \and
YuAn Wang\inst{2} \and Hao Li\inst{2} \and
 Bosheng Wang\inst{2} \and \\
 Xun Sun\inst{2}  \and
 Xin Yu\inst{3}\thanks{Corresponding author: xin.yu@adelaide.edu.au}}

\authorrunning{Du et al.}

\institute{University of Technology Sydney, Australia \and
Baidu Inc., China \and
Australian Institute for Machine Learning,  Adelaide University, Australia \\
}

\maketitle

\begin{figure}

    \centering
    \includegraphics[width=0.99\linewidth]{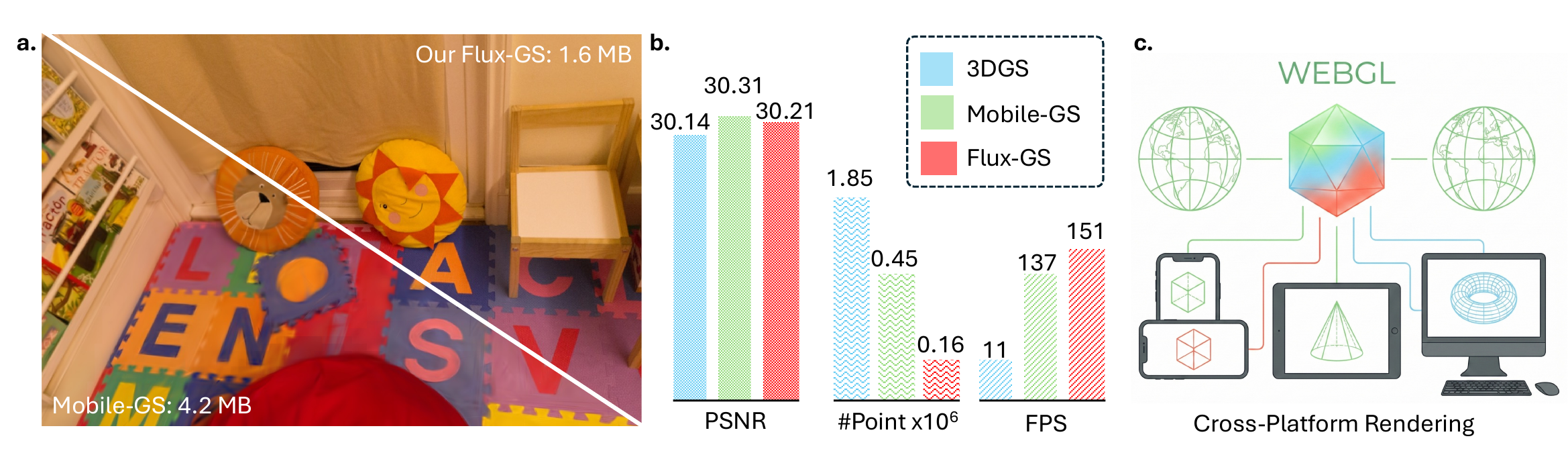}
    
    \caption{ \textbf{ab. Flux-GS} achieves rendering quality comparable to both 3DGS~\cite{kerbl2023gaussiansplatting} and the Mobile-GS~\cite{mobilegs}, while reducing the number of Gaussian primitives and facilitating significantly higher FPS on the mobile with Snapdragon 8 Gen 3 GPU. 
    \textbf{c.} The proposed Flux-GS utilizes WebGL to enable seamless cross-platform rendering. }
    \label{fig:teaser}

\end{figure}

\begin{abstract}
Recent advances in 3D Gaussian Splatting have demonstrated unprecedented success in novel view synthesis.
However, the substantial inference and storage overhead driven by high-order Spherical Harmonics (SH) are primary bottlenecks for mobile platforms. 
In this paper, we present Flux-GS, a real-time Gaussian Splatting method designed to achieve high-fidelity rendering with significantly reduced overhead for resource-constrained mobile platforms. 
We first propose a Monte Carlo Specular Energy Aggregator, sampling third-order radiance residuals and aggregating specular energy into a compact latent space.
In this way, our method effectively preserves visually salient lighting features in lower-order bands without expensive distillation or pre-training. 
To mitigate the high-frequency details lost during compression, we introduce an Attribute-Conditioned SH Enhancement module. 
This module predicts Gaussian-aware offsets based on intrinsic Gaussian attributes, which enhance the first-order SH representation prior to inference, without extra inference costs.
Furthermore, the original single-view gradient-based densification is prone to producing excessive Gaussians and overfitting to a certain view.
We address these limitations by proposing a Multi-view Alpha-based Densification and Pruning strategy.
By leveraging multi-view guidance, we ensure multi-view structure consistency and the precise removal of redundant primitives.
Extensive experiments demonstrate that Flux-GS achieves substantial parameter reduction while maintaining competitive visual quality, offering a robust and scalable solution for real-time mobile rendering.
Code: \textcolor{magenta}{\href{https://xiaobiaodu.github.io/flux-gs-project/}{https://xiaobiaodu.github.io/flux-gs-project/}}.
  \keywords{Gaussian Splatting \and Real-Time Rendering \and 3D Vision }
\end{abstract}

\begin{figure}[t]
    \centering
    \includegraphics[width=0.99\linewidth]{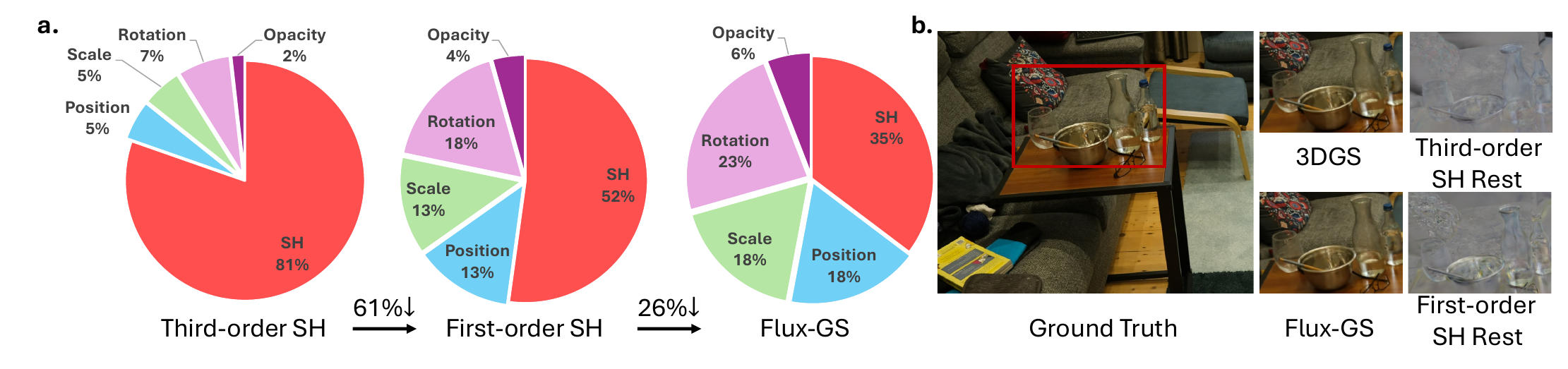}
    
    \caption{\textbf{Gaussian parameter distribution and Spherical Harmonic fidelity analysis. a.} Per Gaussian memory footprint across 3DGS variants. Flux-GS achieves significant compression (61\% and 26\% reductions) by optimizing Spherical Harmonics (SH) coefficients and decoupling SH into the base and view-independent components. 
    \textbf{b.} Qualitative comparison demonstrates that Flux-GS with only first-order SH can render high-fidelity high-frequency details comparable to 3DGS.}
    \label{fig:teaser_sh}
  
\end{figure}

\section{Introduction}
3D scene reconstruction~\cite{yifan2019differentiable, npc} and novel view synthesis~\cite{nerfstudio,mildenhall2021nerf,chen2023mobilenerf} have been revolutionized by the emergence of 3D Gaussian Splatting (3DGS)~\cite{kerbl2023gaussiansplatting}, which has fostered several applications, like navigation~\cite{du2020learning, du2023object, du2021vtnet}, self-driving~\cite{yan2024streetgs, du2024dreamcar, du20243drealcar}, virtual reality~\cite{vrgs, long2023wonder3d, liu2023one, shen2023anything}, and human body understanding~\cite{guo2024being, guo2025plnet, guo2026beyond}.
3DGS offers high-fidelity, real-time rendering through an explicit and differentiable Gaussian-based representation. By modeling scenes as a collection of millions of anisotropic 3D Gaussians, 3DGS achieves unprecedented visual quality while maintaining high frame rates on desktop-class hardware. 
However, the deployment of 3DGS on mobile platforms remains a formidable challenge.

In our experiments, we found two primary bottlenecks that hinder the efficient deployment of 3DGS on mobile devices.
(1) \textbf{Third-order Spherical Harmonics}: 
Previous works~\cite{hanson2024speedy, lee2024compact, zhang2025gaussianspa, liu2025maskgaussian, hou2024sort} typically employ the third-order Spherical Harmonics (SH) to capture complex and view-dependent radiance.
Typically, the third-order SH coefficients require 48 floating-point parameters per Gaussian primitive.  When it scales to millions of Gaussians, it leads to prohibitive storage requirements and memory bandwidth consumption during the rasterization phase.
(2)\textbf{ Redundant Gaussian Points}:
Light-weight GS variants, like Mobile-GS~\cite{mobilegs} and MEGS2~\cite{chen2025MEGS2}, typically utilize the traditional single-view gradient-based densification~\cite{kerbl2023gaussiansplatting}.
This approach aggressively produces an excessive number of primitives and is prone to overfitting. The resulting high primitive count increases training time, storage footprints, and inference latency, making it unsuitable for edge devices.

In this work, we present Flux-GS, a mobile real-time Gaussian Splatting method designed to deliver high-fidelity rendering with significantly fewer parameters. 
As illustrated in Fig.~\ref{fig:teaser}, our proposed Flux-GS achieves rendering quality comparable to existing baselines while utilizing fewer Gaussian primitives, thereby outperforming Mobile-GS~\cite{mobilegs} in inference speed. 
Our framework is built upon two core technical innovations:
(1) \textbf{Monte Carlo Specular Energy Aggregator:} As shown in Fig.~\ref{fig:teaser_sh}, our approach is motivated by the observation that substantial radiance information can be compressed into a lower-order subspace without compromising view-dependent specular features.
 We propose a Monte Carlo Specular Energy Aggregator that aggregates high-order specular energy into a compact representation, effectively preserving the complex lighting characteristics.
Furthermore, we decouple SH into the base and view-independent components, allowing for a substantial reduction in SH memory overhead compared to traditional third-order baselines without the need for expensive offline distillation or pre-training.
To recover the fine detail lost during this compression, we introduce an Attribute-Conditioned SH Enhancement module. This module utilizes a lightweight Multi-Layer Perceptron (MLP) to model contextual offsets based on intrinsic Gaussian properties. More importantly, these offsets are statically baked into the explicit Gaussian parameters prior to inference, ensuring that this enhanced view-dependent modeling does not introduce additional computational overhead during inference.
(2) \textbf{Multi-view Alpha-based Densification and Pruning strategy: }  We further address the inefficiency inherent in the standard single-view gradient-based densification strategy, which often lacks multi-view structure consistency. We propose a Multi-view Alpha-based Densification and Pruning strategy that employs stratified camera sampling and alpha-weighted error accumulation across multiple viewpoints. 
With multi-view guidance, this strategy ensures a more compact representation by accurately pruning redundant primitives while focusing densification on regions with high multi-view reconstruction error.

Extensive experiments demonstrate that Flux-GS offers a robust and scalable solution for mobile real-time rendering. By achieving powerful first-order SH representation while maintaining competitive visual quality, our framework paves the way for the deployment of high-fidelity 3DGS on mobile devices.

\section{Related work}

\noindent\textbf{Gaussian Splatting:}
3D Gaussian Splatting (3DGS)~\cite{kerbl2023gaussiansplatting} has motivated extensive works~\cite{radl2024stopthepop, feng2025flashgs,hamdi2024ges, sun2025stochastic, hahlbohm2025efficient} into accelerating training. For instance, Mini-Splatting~\cite{mini-splatting} improves the densification process via an importance-driven process to explicitly limit the primitive count.
3DGS-MCMC~\cite{mcmc} formulates the training dynamics as a Markov Chain Monte Carlo process, while Revisiting Densification~\cite{rota2024revising} proposes a loss-driven densification scheme. 
MVGS~\cite{du2024mvgs} is the first to propose multi-view regulated training for smooth Gaussian Splatting optimization, which substantially improves multi-view consistency.
Taming 3DGS~\cite{mallick2024taming} enforces a strict user-defined Gaussian budget during optimization.
 Despite these advancements, these variants often remain unsuitable and cost-ineffective for mobile deployment due to their storage footprints and insufficient inference throughput on edge hardware.

\noindent\textbf{Gaussian Pruning:}
The inherent redundancy in 3D Gaussian representations~\cite{niemeyer2025radsplat, 3dgsdr, 2dgs} has motivated recent work on pruning Gaussians to enhance rendering efficiency~\cite{fan2024lightgaussian,mini-splatting, pixelgs, pcgs}. 
Current methods typically rely on importance scores to remove insignificant Gaussians, often calculated by accumulating ray contributions or combining opacity with gradient information~\cite{liu2024compgs, ali2025elmgs}. For instance, EAGLES~\cite{girish2024eagles} evaluates importance via transmittance, and PUP 3DGS~\cite{hanson2025pup} utilizes a Hessian-based sensitivity analysis.
While Mobile-GS~\cite{mobilegs} provides an opacity- and scale-based pruning method, it often fails to preserve fine-grained geometric details.
Fast-GS~\cite{ren2026fastgs} aims to reduce training time with Gaussian compactness.
Moreover, most contemporary approaches still rely on single-view gradient-based densification, which tends to produce an excessive number of Gaussians. In this work, we present Flux-GS, featuring a novel multi-view alpha-based densification and pruning strategy that significantly minimizes the primitive budget while maintaining competitive visual performance.

\noindent\textbf{Gaussian Compression:}
Early approaches adopted traditional compression strategies, including scalar and vector quantization (VQ)~\cite{lee2024compact,papantonakis2024reducing,xie2024mesongs, omg} and entropy coding~\cite{chen2024hac,chen2025hac++}, to decrease storage overhead.
VQ-based methods are highly effective due to the inherent redundancy across Gaussian attributes, enabling compact encoding. 
Recent advancements have increasingly focused on structured representations, combining anchor-based structure~\cite{lu2024scaffold,wang2024contextgs,liu2024compgs} and factorization techniques~\cite{f-3dgs} with grid-based representations~\cite{chen2024hac, ren2024octree}. 
Subsequent research~\cite{chen2024hac,chen2025hac++,wang2024contextgs, li2026gscodec} further proposes contextual modeling to improve compression efficiency.
Despite their strong compression performance, these methods rely on per-view processing with multiple MLP forward passes, resulting in considerable rendering latency.
More recent work leverages neural field architectures to find the local continuity among neighboring Gaussians. 
Compact-3DGS~\cite{lee2024compact} models view-dependent color, while LocoGS~\cite{shin2025locality} encodes all Gaussian primitives, not view-independent information. 
Mobile-GS~\cite{mobilegs} proposes a first-order SH distillation technique to transfer high-frequency information from a high-order teacher model. However,  this distillation process leads to an extremely heavy training burden. 
In this work, we propose Flux-GS, featuring a Monte Carlo Specular Energy Aggregator. Unlike prior approaches, our method compresses third-order SH energy into a low-order representation. It eliminates the need for costly distillation, significantly reducing overhead while maintaining high-fidelity rendering.

\section{Methodology}
Vanilla 3D Gaussian Splatting (3DGS) leverages explicit 3D Gaussians with volume rendering to represent a whole scene or object. Each 3D Gaussian \( \mathcal{G}_i \) is defined by a set of geometry and appearance parameters: \( (\mathbf{\mu}_i, s_i, r_i, o_i, c_i) \).
Specifically, $\mathbf{\mu}_i$, $s_i$, and $r_i$ represent spatial position, scale, and rotation, while $o_i$ and $c_i$ denote opacity and spherical harmonic coefficients for appearance modeling.
To capture complex and view-dependent effects, standard 3DGS methods typically utilize third-order spherical harmonics (SH).  
While effective for modeling specular and intricate lighting, this approach incurs significant storage costs, as each Gaussian primitive requires 48 SH coefficients ($16 \text{ coefficients} \times 3 \text{ channels}$). 
As scene complexity grows into the millions of primitives, the resulting storage footprint and memory bandwidth requirements become prohibitive for mobiles.

In this work, we propose Flux-GS, which constrains the Gaussian appearance modeling to first-order spherical harmonics (requiring only 4 coefficients per channel). To ensure rendering fidelity, we propose a Monte Carlo Specular Energy Aggregator method to compress high-order radiance energy into a compact and lower-order subspace without pretraining or distillation.
By bridging the gap between high-fidelity radiance modeling and hardware-efficient representation, Flux-GS enables high-quality novel view synthesis on edge devices.

\begin{figure}[t!]
    \centering
    \includegraphics[width=0.99\linewidth]{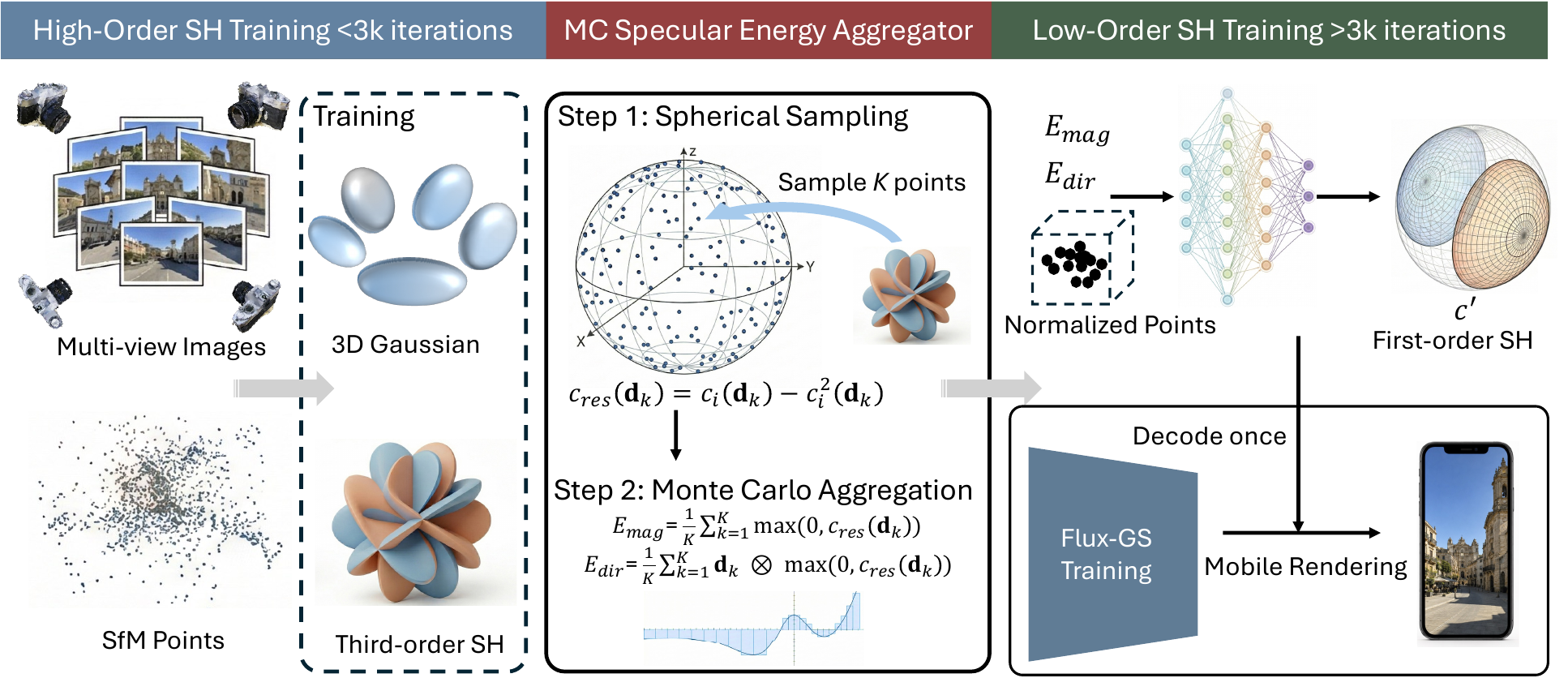}
    
    \caption{\textbf{Overview of the Flux-GS framework.} Our method optimizes third-order SH for the initial 3k iterations, then transitions via Monte Carlo Specular Energy Aggregator for high-frequency representation. With the first-order direction moments inherited from the original third-order SH, we leverage a neural network to aggregate these latents into first-order SH $c'$ for rendering.  During inference, the model requires only once decoding to obtain first-order SH, significantly reducing storage.
   }
    \label{fig:method}
   
\end{figure}

\subsection{Monte Carlo Specular Energy Aggregator}
\textcolor{teal}{$\blacktriangleright$~\textbf{``Insight 1":} Monte Carlo directional moments largely match 1st-order SH.}

\noindent High-order SH polynomials in 3DGS are inherently inefficient for representing sharp, highly localized specular lobes due to their dense and global nature. Fitting a sparse specular peak requires numerous high-order SH merely to cancel out non-specular ringing artifacts. We observe that view-dependent energy is highly sparse in the angular domain, yet its directional phase is critical for accurate reconstruction. Rather than aggregating absolute photometric residuals, which destroys this vital angular gradient, we propose to extract the third-order directional moments of the specular residual via Monte Carlo projection.  It projects the extracted directional energy into the first-order SH space. Consequently, it allows us to discard the redundant high-order SH while explicitly preserving both the magnitude and spatial direction of the view-dependent signal.

Consider an RGB-valued directional function $c_i(\boldsymbol{\omega}) = \sum_{\ell=0}^{L_o}\sum_{m=-\ell}^{\ell} c_{\ell m} Y_{\ell m}(\boldsymbol{\omega})$ defined over the sphere $\mathbb{S}^2$. 
This function is initially represented by spherical harmonics of an original degree $L_o$, with a coefficient tensor: $c \in \mathbb{R}^{N \times (L_o+1)^2 \times 3}$ where $N$ represents the number of Gaussian primitives. Our objective is to compress this high-fidelity representation into a lower-order SH space of degree $L_t$ (where $L_t \leq L_o$), resulting in a reduced coefficient set: $c' \in \mathbb{R}^{N \times (L_t+1)^2 \times 3}$.
For each primitive $i$, we formulate this as a least-squares optimization problem:
\begin{equation}
c'^* = \arg\min_{c'} \int_{\mathbb{S}^2} \left| \left|\ c_i(\omega)  - \sum_{\ell=0}^{L_t} \sum_{m=-\ell}^{\ell} c'_{\ell m} Y_{\ell m} (\boldsymbol{\omega}) \right|\right|_2^2 d\boldsymbol{\omega},
\end{equation}
where $Y_{\ell m}(\boldsymbol{\omega})$ denotes a spherical harmonic basis function. 
To obtain a low-order representation, the optimal projection coefficients $c'_{n,k}$ that minimize the reconstruction error are given analytically by the inner product:
\begin{equation}
c'_{i,\ell m} = \langle c_i(\boldsymbol{\omega}) , Y_{\ell m} \rangle=\int_{\mathbb{S}^2}
c_i(\boldsymbol{\omega}) Y_{\ell m}(\boldsymbol{\omega}) 
d\boldsymbol{\omega},
\label{eq:project}
\end{equation}
where it effectively concentrates the signal energy into the lower-order bands.
However, this analytical projection directly discards critical view-dependent specular details.
To retain these high-frequency characteristics without incurring the prohibitive high-order SH, we aim to compress the discarded specular information into a compact latent representation.
A naive approach is to aggregate the raw SH polynomials over the spherical domain. However, this is mathematically futile due to the orthogonality of the SH function. The integral of any purely high-order SH polynomial ($l \ge 1$) over the sphere evaluates to strictly zero. 
Recognizing that the true perceptual impact of a specularity is fundamentally characterized by its photometric energy magnitude and dominant direction, we instead propose sampling the high-frequency residual on a uniform sphere to extract its first-order directional moments.

\noindent\textbf{Uniform Spherical Sampling:}
To generate a set of directional vectors $\mathbf{D} = \{\mathbf{d}_k\}_{k=1}^K$, we uniformly sample $K$ points on the unit sphere $\mathbb{S}^2$ using spherical coordinates. For each sample $k$, we draw two independent random variables $\theta_k$ and $\phi_k$ respectively representing the colatitude and longitude:
\begin{equation}
   \quad \phi_k \sim \mathcal{U}(0, 2\pi), \quad  \theta_k = \arccos(1 - 2\xi_k), \quad \xi_k \sim \mathcal{U}(0, 1),
\end{equation}
where $\mathcal{U}$ denotes the uniform sampling.
These spherical coordinates are subsequently transformed into the Cartesian domain to obtain the unit direction vectors 
$
\mathbf{d}_k = \begin{bmatrix}
\sin \theta_k \cos \phi_k, ~
\sin \theta_k \sin \phi_k, ~
\cos \theta_k
\end{bmatrix}
$
where each $\mathbf{d}_k$ represents a unit vector such that $\|\mathbf{d}_k\|_2 = 1$. This collection of vectors $\mathbf{D} \in \mathbb{R}^{K \times 3}$ serves as the set of incident directions for subsequent spherical harmonic compression.

\noindent\textbf{Energy Aggregation:}
Subsequently, we evaluate the photometric residual between the third-order SH radiance $c_{i}(\mathbf{d}_k)$ and the second-order SH radiance $c_{i}^{2}(\mathbf{d}_k)$ via $c_{res}(\mathbf{d}_k) = c_{i}(\mathbf{d}_k) - c_{i}^{2}(\mathbf{d}_k)$. By aggregating these residuals, we formulate a compact, high-frequency dynamic energy and direction representation:
\begin{equation}
E_{mag} = \frac{1}{K} \sum_{k=1}^{K} \max(0, c_{res}(\mathbf{d}_k)), \quad E_{dir} = \frac{1}{K} \sum_{k=1}^{K} \mathbf{d}_k \otimes \max(0, c_{res}(\mathbf{d}_k))
\label{eq:sh_hq}
\end{equation}
where $\otimes$ represents the outer product. $E_{mag}$ and $E_{dir}$ effectively encapsulate the overall energy and direction of the view-dependent specularities for each Gaussian primitive, compressing the bulky high-order parameters into a highly lightweight low-dimensional descriptor.
To strictly minimize memory overhead, we only store these compact latents $E_{mag}$ and $E_{dir}$. We then employ a neural network to learn a non-linear mapping from this aggregated high-frequency latent space to the first-order SH parameters, conditioned on the normalized Gaussian position:
\begin{equation}
    c' = \Psi(E_{mag},E_{dir}, f(\hat{\mu})), \quad
f(\hat{\mu}) = 
\begin{cases} 
\hat{\mu} & \text{if } \|\hat{\mu}\|_2 \leq 1, \\
\left( 2 - \frac{1}{\|\hat{\mu}\|_2} \right) \frac{\hat{\mu}}{\|\hat{\mu}\|_2} & \text{if } \|\hat{\mu}\|_2 > 1,
\end{cases}
\end{equation}
where $\Psi$ is a Multi-Layer Perceptron (MLP), and the result $c' \in \mathbb{R}^{N \times (L_t+1)^2 \times 3}$ represents the first-order SH coefficient ($L_t=1$). 
We normalize the Gaussian position $\mu$ to reduce the impact of spatial outliers for better convergence.
By performing this mapping, Flux-GS can dynamically adjust the Gaussian geometry and low-order appearance simultaneously, ensuring that the restricted SH capacity is allocated to the most visually salient scene features. 
In particular, our approach eliminates the need for expensive distillation while achieving significant compression in SH storage memory overhead compared to third-order baselines. More importantly, we only decode first-order SH coefficients once before inference, so it would not affect inference overhead.

\noindent\textbf{Gaussian Attribute Awareness:}
Although the above process preserves the majority of radiance energy in an $L^2$ sense, projecting higher-order bands inevitably degrades overall color fidelity and scene structure.
We observe that this discarded residual energy is highly correlated with intrinsic Gaussian attributes, such as spatial position and anisotropic scale.
This motivates us to design a lightweight geometry-conditioned residual module to contextually refine the base representation and preserve the efficiency of the first-order SH.
Specifically, we introduce an Attribute-Conditioned SH Enhancement module parameterized by a lightweight MLP, denoted as $\Phi$, to predict a residual offset $\Delta c$ for the first-order SH coefficients $c'$. The network aggregates the geometric and photometric properties by concatenating the normalized first-order SH $\hat{c}'$, opacity $o$, normalized scales $\hat{s}$, spatial position $\mu$, and rotation quaternion $r$. The enhanced SH coefficients $c^{out}$ are formulated as:
\begin{equation}
    \Delta c_i = \Phi([\mathbf{\mu}_i, \hat{s}_i, r_i, o_i, \hat{c}'_i ]), \quad c^{out}_i = c'_i + \Delta c_i,
\end{equation}
where the final linear layer of $\Phi$ is strictly zero-initialized to ensure optimization stability, allowing the model to naturally fall back to the standard 3DGS representation at the beginning of training and smoothly learn the residuals. Crucially, because $\Phi$ relies strictly on intrinsic Gaussian attributes and is entirely independent of the camera viewing direction, the residual $\Delta c$ only needs to be decoded once for subsequent inference. Prior to rendering, the SH offset is pre-computed and statically baked into the explicit Gaussian parameters. This design guarantees that our contextual modeling facilitates higher performance while introducing absolutely zero computational overhead during inference.

\noindent\textbf{Training Curriculum:} 
Initially, we optimize 3D Gaussians utilizing full third-order SH coefficients for the first 3k iterations to establish a high-frequency representation. Subsequently, we aggregate these third-order coefficients down to a first-order SH representation via our proposed Monte Carlo Specular Energy Aggregator. After this mapping step, we initialize the Attribute-Conditioned SH Enhancement module to continually refine the low-order representation.
Note that the compression from third-order to first-order is executed only once. After that, we only maintain and update the lightweight first-order parameters.

\begin{figure}[t]
    \centering
    \includegraphics[width=0.99\linewidth]{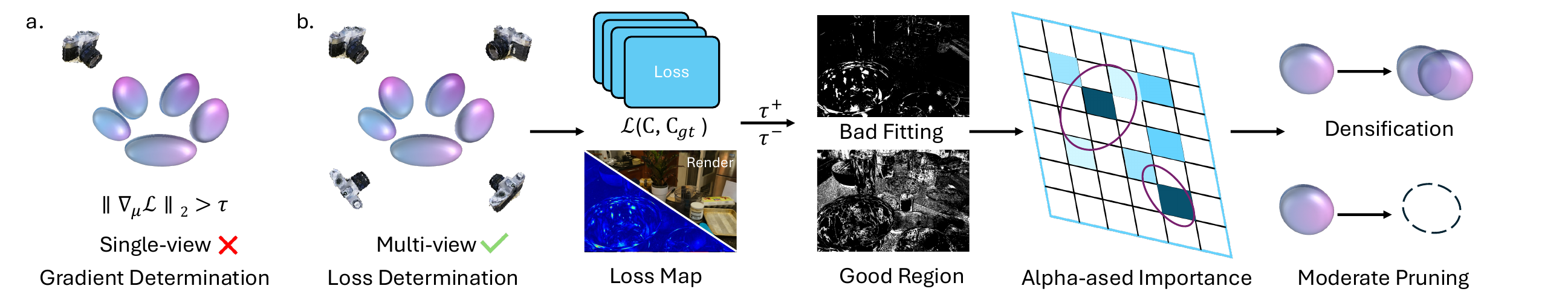}
    
    \caption{\textbf{Illustration of our proposed Multi-view Alpha-based Densification and Pruning strategy.} \textbf{a.} Traditional 3DGS leverages the single-view Gaussian position gradient determinist. \textbf{b.} We propose to employ a multi-view loss-driven mechanism, coupled with alpha-based Gaussian evaluation to densify more Gaussians in the bad-reconstructed regions and moderately prune Gaussians for well-reconstructed areas. }
    \label{fig:densify}
  
\end{figure}

\subsection{Multi-view Alpha-based Densification and Pruning}
\textcolor{teal}{$\blacktriangleright$~\textbf{``Insight 2":} Multi-view Gaussian variation ensures multi-view consistency.}

\noindent  To achieve a compact 3D Gaussian structure, we found that an accurate densification and pruning strategy is crucial. We observed previous methods~\cite{kerbl2023gaussiansplatting, mini-splatting} using gradient-based or importance-based strategies in the single-view criterion to determine Gaussians' densification and pruning. However, the single-view criteria lack a holistic understanding of the scene, as they account for geometry and appearance from only a single view. Therefore, we propose a multi-view alpha-based densification and pruning strategy in Fig.~\ref{fig:densify} to constrain the number of Gaussians based on multi-view evaluation so that ensure multi-view consistency.

\noindent\textbf{Stratified Camera Sampling:}
To ensure comprehensive geometric coverage and alleviate view redundancy within the training camera set, we adopt a \textit{Stratified Camera Sampling} strategy. We partition the 3D scene into angular bins defined with respect to a focal point, enabling the selection of representative views that maximize view variation and angular dispersion.
Prior to binning, we estimate the center of the scene, denoted as $\mathbf{x}^*$. Given $N_{cam}$ cameras with the centers $\{\mathbf{t}_i\}_{i=1}^{N_{cam}}$ and viewing directions $\{\mathbf{v}_i\}_{i=1}^N$, we compute the point that minimizes the total squared distance to all camera rays. Let $P_i = I - \mathbf{v}_i \mathbf{v}_i^\top$ be the projection matrix onto the subspace orthogonal to the $i$-th ray, where $I$ is an identity matrix. The scene center $\mathbf{x}^*$ is obtained by solving the linear least-squares system via:
$
    \left( \sum_{i=1}^{N_{cam}} P_i \right) \mathbf{x}^* 
= 
\sum_{i=1}^{N_{cam}} P_i \mathbf{t}_i,
$
where each camera position is transformed into spherical coordinates $(\theta, \phi, \rho)$. $\theta = \mathrm{arctan2}(r_y, r_x)$, $\phi = \arcsin(r_z / \|\mathbf{r}_i\|)$, and $\rho = \|\mathbf{r}_i\|$, with $\mathbf{r}_i = \mathbf{t}_i - \mathbf{x}^*$. The angular space is discretized into $M_\theta$ azimuth and $M_\phi$ elevation bins. Cameras are grouped according to their bin indices, and one representative is uniformly sampled from each non-empty bin. Finally, to satisfy a computational budget $K$, the selected candidates are randomly shuffled and truncated, yielding a camera subset $\mathcal{C}$ with broad and uniform angular coverage.

\noindent\textbf{Photometric Loss Guidance:}
To effectively optimize the Gaussian Splatting representation, we employ a multi-view consistency-driven strategy. This strategy identifies regions of high and low reconstruction error across multiple viewpoints and uses the accumulated contribution of each Gaussian to guide geometry and appearance refinement. For each viewpoint in a curated camera set $\mathcal{C}$, we render the current Gaussian primitives and compute a pixel-wise photometric loss. Following standard 3DGS, we utilize the loss with a hybrid objective combining $L_1$ distance and Structural Similarity (SSIM) through:
\begin{equation}
\mathcal{L} = (1 - \lambda) \mathcal{L}_1(\mathbf{C}, \mathbf{C}_\text{gt}) + \lambda \mathcal{L}_{\text{DSSIM}}(\mathbf{C}, \mathbf{C}_\text{gt}),
\label{eq_loss}
\end{equation}
where $\mathbf{C}$ and $\mathbf{C}_\text{gt}$ represent the rendered image and Ground Truth. $\lambda$ controls the weight for $\mathcal{L}_1$ and $\mathcal{L}_{\text{DSSIM}}$.
Therefore, we can derive a binary Metric Map $M \in \{0, 1\}^{H \times W}$ to identify areas of poor and high-quality reconstruction through:
\begin{equation}
        M_{u,v}^{+} = \mathbb{I}\left( \mathcal{L}_{u,v}> \tau^{+} \right), \quad 
    M_{u,v}^{-}= \mathbb{I}\left( \mathcal{L}_{u,v} < \tau^{-}  \right),
\end{equation}
where $\mathbb{I}(\cdot)$ is the indicator function and $\tau$ is the error threshold.
A pixel $(u, v)$ is flagged ($M_{u,v}^+ = 1$) if its error exceeds a predefined threshold $\tau^+$, marking it as a region of poor reconstruction. On the contrary, $M_{u,v}^-=1$ means that the region is well reconstructed, needed to prune redundant Gaussian points.

\noindent\textbf{Multi-view Alpha-Weighted Error Accumulation:}
To map 2D image-space errors back to 3D Gaussian primitives, we customize the CUDA kernel for the rendering pipeline. For each Gaussian $i$, we compute the importance score $S_i^+$ and pruning score $S_i^-$ by summing its alpha contributions $\alpha_{i}$ across all flagged pixels in the metric map by:
\begin{equation}
    S_i^+ = \sum_{c=0}^\mathcal{C} \sum_{(u,v) \in \Omega } \alpha_{i} \cdot M_{uv}^{c,+}, \quad
        S_i^- = \sum_{c=0}^\mathcal{C} \sum_{(u,v) \in \Omega } \mathcal{L}_{uv} \cdot \alpha_{i} \cdot M_{uv}^{c,-},
\end{equation}
where  $\alpha_i = o_i \exp\left(-\frac{1}{2}\Delta x_i^T \Sigma_i^{-1} \Delta x_i\right)$ represents Gaussian alpha, indicating the Gaussian contribution to the pixel color. $\Omega \subset \mathbb{Z}^2$ denote the discrete image pixel domain. We aggregate these local metrics across the sampled views to compute these two primary scores. This multi-view aggregation effectively differentiates Gaussians that are important or redundant, allowing for more accurate geometry modification. By averaging across the sampling views, our Flux-GS avoids over-fitting to single-view occlusions or transient artifacts.
By using the alpha-based metric rather than the conventional gradient metric~\cite{kerbl2023gaussiansplatting}, we ensure that only the Gaussians with high visibility and contributing to the error are modified.

\noindent\textbf{Accurate Densification:}
The vanilla 3DGS employs a gradient-based strategy to identify underfitting Gaussians. We define the vanilla clone mask $\mathcal{M}_{clone}^{base}$ and split mask $\mathcal{M}_{split}^{base}$ based on standard positional gradient threshold $\|\nabla_{\boldsymbol{\mu}_i} \mathcal{L}\|_2 > \tau$. 
However, relying solely on single-view positional gradients frequently results in excessive and redundant Gaussian densification.
To mitigate this issue, we integrate our computed importance score into the traditional 3DGS gradient-based densification strategy. Our multi-view consistency-guided masks are formed by intersecting these base criteria with an importance quantile filter:
\begin{equation}
    \mathcal{M}_{clone} = \mathcal{M}_{clone}^{base} \land \left( S^+  > Q_\tau^{+}(S^+)   \right), ~
    \mathcal{M}_{split} = \mathcal{M}_{split}^{base} \land \left( S^+  >  Q_\tau^{+}(S^+) \right),
\end{equation}
where $ Q_\tau^{+}$ denotes the quantile of the importance score distribution. This ensures that Gaussians are only cloned or split if they consistently contribute to high-error regions across multiple views. Consequently, these refined masks strictly govern the cloning and splitting operations, boosting Gaussians' fitting capability while tightly controlling the primitive count.

\noindent\textbf{Moderate Pruning:}
Standard 3DGS defines a threshold $o_{min}$ to eliminate Gaussian points with low opacity, leading to aggressive pruning and performance slump. To achieve moderate pruning, we introduce the pruning score $S^-$ based on multi-view and loss-driven strategy. A larger $S^{-}$ indicates the Gaussian that impacts more to low-error pixel regions, effectively flagging it as an artifact or redundant point. Our final pruning mask $\mathcal{M}_{prune}$ identifies Gaussians that fall below the opacity threshold $o_{min}$ and exceed the pruning score quantile $Q_\tau^{-}$ via:
\begin{equation}
    \mathcal{M}_{prune} = (o_i < o_{min}) \land \left( S^- >  Q_\tau^{-}(S^-) \right),
\end{equation}
where we utilize $\mathcal{M}_{prune}$ to cull non-contributing or artifact-inducing Gaussians while preserving essential scene geometry. In this way, we achieve moderate pruning while maintaining overall reconstruction fidelity.

\subsection{Implementation}

\noindent \textbf{Training Details}: 
We train our method with 30k iterations, following the vanilla 3DGS~\cite{kerbl2023gaussiansplatting}, without the need for pretraining or distillation stages like Mobile-GS~\cite{mobilegs}.  To train our proposed Flux-GS, we employ a hybrid photometric loss function consistent with the original 3DGS~\cite{kerbl2023gaussiansplatting}. 
As shown in Eq.~\ref{eq_loss}, we employ this loss to jointly optimize the Gaussian parameters and the compressed SH features.
More training details can be found in the supplementary material for reproduction.

\noindent \textbf{Deployment on Mobiles}: 
To enable cross-platform real-time rendering, we introduce a WebGL-based rendering framework for our Flux-GS. Unlike traditional CUDA-dependent pipelines, our approach leverages the ubiquitous WebGL API to achieve real-time rendering directly within browsers, without external dependencies. 
To address the non-commutative nature of alpha-compositing, we implement a decoupled sorting strategy. 
By offloading the depth-sorting of millions of Gaussians to an asynchronous WebWorker and using the CPU for the visibility order update, we can break through the 120 display FPS limit in the rendering loop. 
This framework significantly reduces the barrier for navigable 3D environments, providing a cross-platform solution for real-time rendering.

\section{Difference with Mobile-GS}

\noindent\textbf{Energy Aggregation vs. Distillation-based SH:} 
Flux-GS significantly improves training efficiency by replacing the teacher-student distillation of Mobile-GS~\cite{mobilegs} with a Monte Carlo Specular Energy Aggregator. This method maps third-order radiance energy into a compact low-order subspace, preserving high-frequency signal without requiring a distillation process.

\noindent\textbf{Zero-Cost SH Enhancement:} To recover the high-frequency details lost during SH compression, previous Mobile-GS utilizes a neural view-dependent enhancement strategy that conditions on viewing direction. While effective, this introduces a neural network into the rendering pipeline, significantly hindering rendering efficiency. 
Instead of dynamic computation for view-dependent information, our Flux-GS proposes an Attribute-Conditioned SH Enhancement module to predict view-independent SH offsets that are statically baked into the explicit Gaussian parameters before the inference stage. This ensures that the enhanced view-dependent modeling introduces zero additional computational overhead during rendering, since it is decoded only once before inference, facilitating higher rendering speed.

\noindent\textbf{Multi-view vs. Single-view Consistency:}
Mobile-GS relies on an opacity- and scale-based pruning strategy, alongside traditional single-view gradient-based densification. This approach can lead to an excessive number of Gaussian primitives and potential overfitting to a single viewpoint. Flux-GS addresses these limitations with a Multi-view Alpha-based Densification and Pruning strategy. By leveraging a multi-view alpha-weighted error guidance, it allows for the precise removal of redundant Gaussian primitives, resulting in a more compact representation and ensuring multi-view consistency.

\begin{table*}[t!]
\centering

\caption{\textbf{Comprehensive evaluation on the mobile device with Snapdragon 8 Gen 3 GPU on the Mip-NeRF 360 dataset~\cite{mip-nerf}.}
To facilitate a comprehensive analysis of Flux-GS, scenes are categorized into indoor and outdoor subsets.
\#G denotes the number of Gaussian primitives.
 3DGS* represents the quantized version through Huffman encoding for mobile rendering.
 Mobile-GS* denotes the version of Mobile-GS~\cite{mobilegs} without MLP in the inference stage.
 We compare with this version without MLP for fairness.
The \colorbox{green!30}{best} and \colorbox{lime!30}{second-best} results are highlighted.
}

\scriptsize
\def\arraystretch{1.2}
\resizebox{0.99\linewidth}{!}
{
    \begin{tabular}{ l  |ccccc|ccccc}
    \toprule
    Category& \multicolumn{5}{c|}{Indoor}  &  \multicolumn{5}{c}{Outdoor}  \\
    
    Method \& Metrics& PSNR$\uparrow$ & \#G $\times 10^6$$\downarrow$& Storage (MB)$\downarrow$&FPS$\uparrow$&Train (min)$\downarrow$& PSNR$\uparrow$ &  \#G $\times 10^6$$\downarrow$& Storage (MB)$\downarrow$&FPS$\uparrow$ &Train (min)$\downarrow$\\
      \midrule   \midrule 

    3DGS~\cite{kerbl2023gaussiansplatting} &30.41&1.45& 478& -&27&24.61& 3.14& 1361&  -&   36\\

    \midrule

    3DGS* &30.04&1.45& 46&  11&27&   24.39& 3.14& 85& 5&36\\
    
    

    Speedy-Splat~\cite{hanson2024speedy}  &30.11&0.37& 64& 21&\colorbox{lime!30}{14}&  24.41& 0.56& 88& 14&\colorbox{lime!30}{13}\\

    C3DGS~\cite{lee2024compact}  & 30.01& 0.75& 21 & 18&31&  24.38&  0.91&  34&  13&45\\

    LocoGS-S~\cite{shin2025locality}  &  30.08&0.82&  6.1&  13&46&  24.31& 1.41&  10&   19& 61\\

    Mobile-GS~\cite{mobilegs}    & \colorbox{green!30}{30.37}& 0.38 & 3.5  &   131  & 86 & \colorbox{green!30}{24.51} &\colorbox{lime!30}{0.58} &5.5  &114 & 136 \\  

        Mobile-GS*   & 29.58& \colorbox{lime!30}{0.38} & \colorbox{lime!30}{3.4}  &   \colorbox{lime!30}{142}& 64& 23.74&\colorbox{lime!30}{0.58} &\colorbox{lime!30}{5.4}  &\colorbox{lime!30}{128}& 114\\  \midrule

    Flux-GS (\textbf{Ours})    & \colorbox{lime!30}{30.22} & \colorbox{green!30}{0.22} & \colorbox{green!30}{2.1 }& \colorbox{green!30}{147}  & \colorbox{green!30}{11} & \colorbox{lime!30}{24.45} &\colorbox{green!30}{0.48}  &  \colorbox{green!30}{4.6} &\colorbox{green!30}{132} & \colorbox{green!30}{11} \\  \bottomrule

    \end{tabular}
}

\label{table:360}

\end{table*}





    
    



     

   
    

\section{Experiments}

\subsection{Hardware Platform}
We conduct all mobile-centric rendering experiments on a commercial smartphone equipped with the Qualcomm Snapdragon 8 Gen 3 GPU. 
To ensure fair and reproducible evaluation, we measure rendering performance utilizing an offscreen benchmarking protocol, which breaks the screen display 120 FPS limit and UI overhead. Specifically, frames are rendered to an offscreen frame buffer, and the average FPS is computed over multiple consecutive runs after a short warm-up period to mitigate thermal and initialization effects.
As for the desktop GPU, we use a single RTX 4090.


\subsection{Qualitative and Quantitative Results}

\noindent\textbf{Quantitative Evaluation and Mobile Performance:}
As detailed in Tables~\ref{table:360} and~\ref{table:tant}, we evaluate Flux-GS against state-of-the-art lightweight Gaussian Splatting variants, including 3DGS~\cite{kerbl2023gaussiansplatting}, Speedy-Splat~\cite{hanson2024speedy}, C3DGS~\cite{lee2024compact}, LocoGS~\cite{shin2025locality}, and Mobile-GS~\cite{mobilegs} across the Mip-NeRF 360~\cite{mip-nerf360}, Tanks and Temples~\cite{tanktemple}, and Deep Blending~\cite{deepblending} datasets. To ensure a fair evaluation of baseline rendering efficiency, we compare against a variant of Mobile-GS that omits the Multi-Layer Perceptron (MLP) during inference, as the MLP introduces significant computational overhead.
Notably, our proposed method achieves a significant breakthrough in the trade-off between rendering fidelity and computational efficiency.  The original 3DGS and its quantized counterpart 3DGS* incur substantial storage costs and exhibit slow rendering speed on the mobile platform. Flux-GS maintains competitive image quality metrics while consistently delivering the fastest FPS across all tested indoor and outdoor environments. Furthermore, by optimizing the Gaussian distribution and utilizing our proposed compressed first-order SH representation, our method achieves the lowest memory footprint among all baselines, compressing entire scenes to very low storage costs. This minimal storage requirement makes it particularly well-suited for resource-constrained edge devices.
Beyond inference efficiency, a key advantage of Flux-GS is its drastically reduced training time. 
Our model achieves convergence several times faster than prior state-of-the-art methods, including Speedy-Splat and Mobile-GS. This accelerated convergence is directly attributed to our proposed multi-view alpha-based densification and pruning strategy, which minimizes the number of Gaussian primitives during optimization.
Flux-GS achieves state-of-the-art rendering efficiency, offering rapid training and minimal storage.

\begin{table*}[t!]
\centering

\caption{\textbf{Quantitative evaluation of state-of-the-art light-weight Gaussian-based methods on the real-world datasets.} We report performance on the Tank\&Temples~\cite{tanktemple} and Deep Blending~\cite{deepblending} datasets.  
 }
 
\scriptsize
\def\arraystretch{1.2}
\resizebox{0.99\linewidth}{!}
{
    \begin{tabular}{ l  |cccccc|cccccc}
    \toprule
    Dataset  & \multicolumn{6}{c|}{Tanks\&Temples}  &  \multicolumn{6}{c}{Deep Blending}  \\
    
    Method \& Metrics& PSNR$\uparrow$ & SSIM$\uparrow$ & LPIPS$\downarrow$  & Storage (MB) $\downarrow$&FPS$\uparrow$ &Train (min)$\downarrow$ & PSNR$\uparrow$ & SSIM$\uparrow$ & LPIPS$\downarrow$  & Storage (MB) $\downarrow$&FPS$\uparrow$ &Train (min)$\downarrow$ \\
      \midrule   \midrule 

    
    


    3DGS~\cite{kerbl2023gaussiansplatting} &23.14 & 0.841&0.183  & 358.7& -&34&29.41&0.903 & 0.243 & 697.3   &  -&   24\\    \midrule

    3DGS* &23.02& 0.824&0.251& 358.7& 7&\colorbox{lime!30}{34}&29.17&0.865& 0.311& 697.3   &  12&   24\\


    
    

    Speedy-Splat~\cite{hanson2024speedy}  &23.08 &0.821 &0.241 & 62.4  & 16&\colorbox{lime!30}{34}&  29.11 & 0.864 &0.309    & 71.2  & 24&\colorbox{lime!30}{16}\\

    C3DGS~\cite{lee2024compact}  & \colorbox{green!30}{23.32} & 0.831 &\colorbox{lime!30}{0.202} & 21.8  & 15&49&  \colorbox{lime!30}{29.73} & 0.900 & 0.258    &  24.7  &  16&35\\

    LocoGS-S~\cite{shin2025locality}  &  23.23 &\colorbox{green!30}{0.837} &\colorbox{green!30}{0.204} &  6.8  &  21&65&  29.76 & \colorbox{lime!30}{0.903} &\colorbox{lime!30}{0.251}    &  7.8  &   14& 47\\

    Mobile-GS~\cite{mobilegs}    & 23.09 & \colorbox{lime!30}{0.831}&0.208  & \colorbox{lime!30}{2.5}  &   124&  82 &\colorbox{green!30}{29.93}&\colorbox{green!30}{0.906} & \colorbox{green!30}{0.243} &4.6  &135&  75 \\  

    Mobile-GS*   & 22.41& 0.825&0.227& \colorbox{lime!30}{2.5}  &   \colorbox{lime!30}{129}&  71&29.14&0.895& 0.281&\colorbox{lime!30}{4.5}  &\colorbox{lime!30}{146}&  63\\  \midrule
        
    Flux-GS (\textbf{Ours})    & \colorbox{lime!30}{23.27} &  0.827  &0.221  & \colorbox{green!30}{2.4}& \colorbox{green!30}{137}& \colorbox{green!30}{11} &\colorbox{lime!30}{29.73}&0.898 & 0.275 &\colorbox{green!30}{1.9} &\colorbox{green!30}{158}& \colorbox{green!30}{10}\\   \bottomrule

    \end{tabular}
}

\label{table:tant}

\end{table*}

\begin{figure}[t!]
    \centering
    \includegraphics[width=0.99\linewidth]{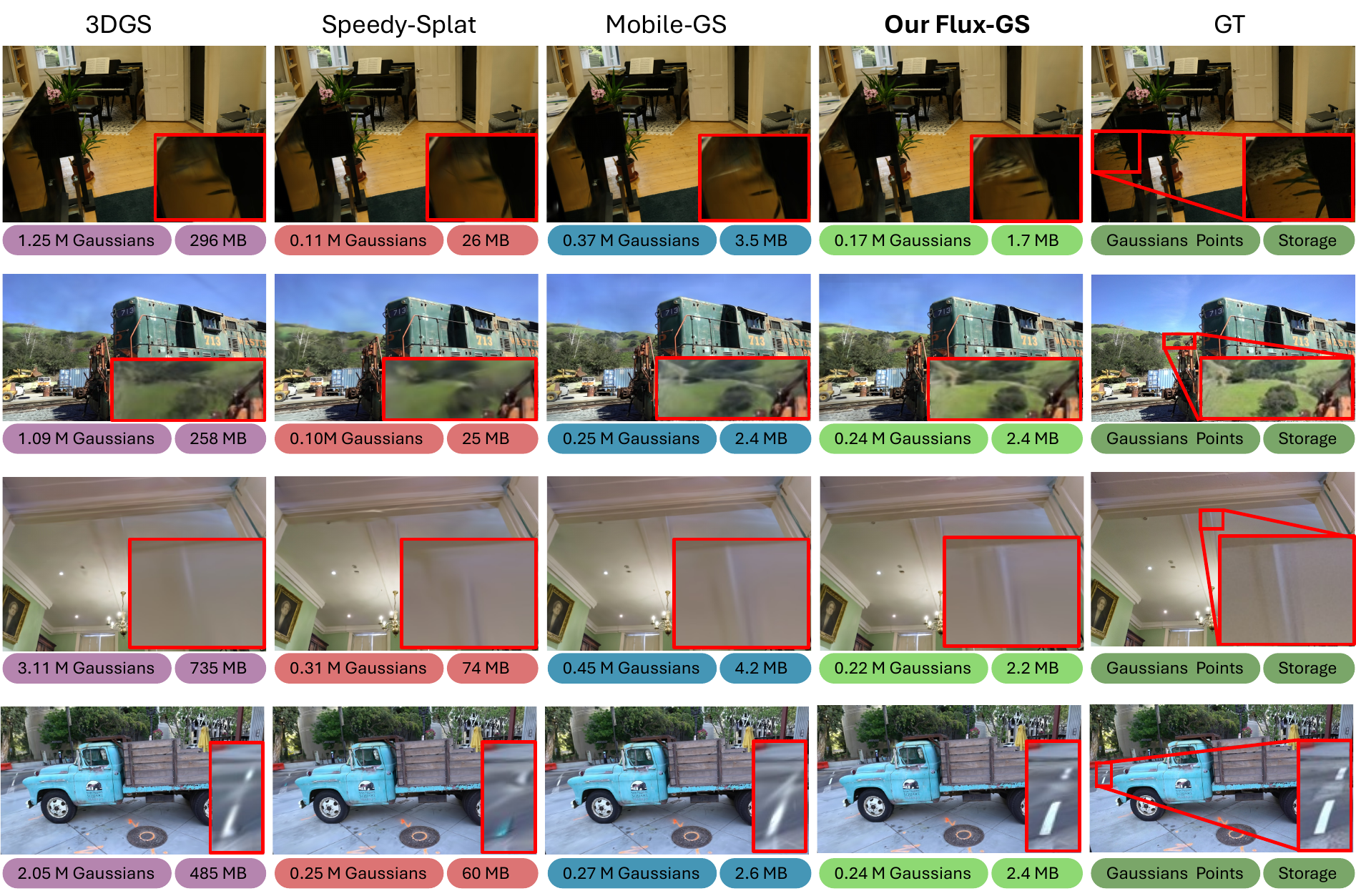}
        
    \caption{\textbf{Qualitative and efficiency comparison with previous state-of-the-art methods.}
We compare rendering quality, Gaussian number, and storage costs across 3DGS, Speedy-Splat, Mobile-GS, and our Flux-GS. 
Zoomed-in regions highlight details and structural consistency for clearer differentiation. 
}
    \label{fig:nvs}
  
\end{figure}

\noindent\textbf{Qualitative Results:}
Fig.~\ref{fig:nvs} provides a qualitative comparison of Flux-GS against advanced methods, including 3DGS~\cite{kerbl2023gaussiansplatting}, Speedy-Splat~\cite{hanson2024speedy}, and Mobile-GS~\cite{mobilegs}. While 3DGS achieves high fidelity, its massive Gaussian count prohibits mobile deployment. Conversely, Speedy-Splat reduces primitives but introduces significant blurring in complex regions. Flux-GS effectively bridges this gap. By optimizing the Gaussian distribution, our method preserves high-frequency details and sharp structural edges, such as mechanical components and textures, with greater clarity than previous methods. Across all scenarios, Flux-GS consistently utilizes fewer primitives to approximate Ground Truth, yielding minimal storage requirements and enabling high-quality mobile rendering.

\begin{table}[t!]
\centering
\caption{\textbf{Ablation study of our Flux-GS on the Mip-NeRF 360 dataset.} 
MC-SEA denotes the proposed Monte Carlo Specular Energy Aggregator. 
$\Delta c$ represents the SH offset predicted by our Attribute-Conditioned SH Enhancement module. 
We also ablate the proposed Multi-view Alpha-based Densification and Pruning.
}

\scriptsize
\def\arraystretch{1.2}
\resizebox{0.89\linewidth}{!}
{
\begin{tabular}{l|cccccc}
\toprule
Method & PSNR $\uparrow$ & Storage (MB) $\downarrow$                     &Peak VRAM$\downarrow$& FPS$\uparrow$                               & \#Points $\times 10^6$ $\downarrow$          & Train (min) $\downarrow$                  \\ \hline
Flux-GS& 27.02   &\colorbox{green!30}{3.5}   &\colorbox{green!30}{388}&\colorbox{green!30}{139} &\colorbox{green!30}{0.36} &  \colorbox{lime!30}{11}\\ \midrule
                  w/o MC-SEA  &                 26.64&                                               \colorbox{green!30}{3.5}   &\colorbox{green!30}{388}&                                              \colorbox{green!30}{139} &                                              \colorbox{green!30}{0.36} &                                           \colorbox{lime!30}{11}\\
 w/o $\Delta c$ & 26.71& \colorbox{green!30}{3.5}   & \colorbox{green!30}{388}& \colorbox{green!30}{139} & \colorbox{green!30}{0.36} & \colorbox{green!30}{9}\\
                  w/o Multi-view densify&                \colorbox{green!30}{27.14}&                                               16&1591&                                              24&                                              \colorbox{lime!30}{1.37}&                                           18\\ 
                  w/o Multi-view prune&                 \colorbox{lime!30}{27.09}&                                               \colorbox{lime!30}{4.8}&\colorbox{lime!30}{416}&                                              \colorbox{lime!30}{119}&                                              0.59&                                           13\\ \bottomrule
\end{tabular}}
\label{table:ablation}

\end{table}

\subsection{Ablation Study}
We analyze the contribution of each component in Flux-GS by progressively disabling these components, including Monte Carlo Specular Energy Aggregator (MC-SEA), enhanced SH offset $\Delta_c$, and Multi-view Alpha-based Densification and Pruning strategy. 
The full model achieves the best trade-off between reconstruction quality and efficiency, attaining competitive PSNR while requiring lower storage costs, fewer Gaussian points, and delivering the fastest FPS on mobile hardware. 
Removing MC-SEA or the SH offset degrades reconstruction quality while maintaining similar efficiency, indicating their positive and complementary contributions.
To ablate the proposed multi-view densification, we adopt the original single-view gradient-based densification~\cite{kerbl2023gaussiansplatting} for replacement.
Disabling multi-view densification substantially increases the number of Gaussians, memory usage, and storage cost.
It significantly reduces FPS, demonstrating the critical role of our proposed multi-view densification for compact Gaussian structure.
Similarly, removing multi-view pruning increases model size and reduces rendering speed, highlighting its importance in eliminating redundant primitives.
Overall, these results demonstrate that our proposed components contribute to achieving an improved trade-off between quality and efficiency, and their combination is essential for high-quality real-time mobile rendering.

\begin{figure}[t!]
    \centering
    \includegraphics[width=0.99\linewidth]{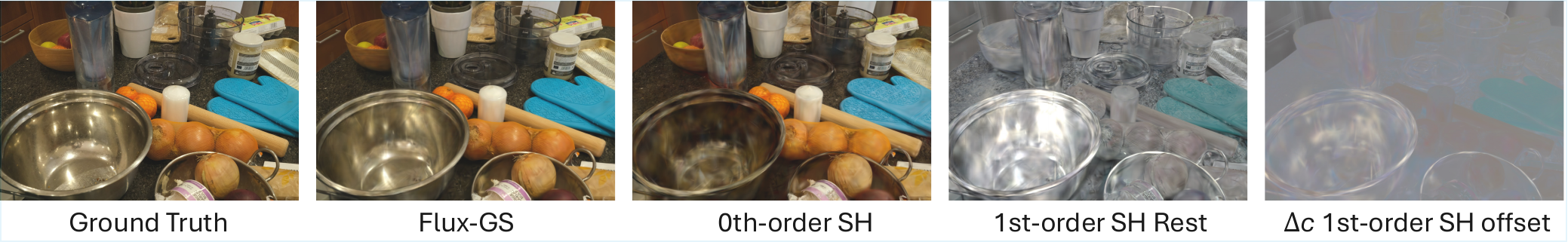}
        
    \caption{\textbf{Decomposition of the Spherical Harmonic components.} 
    The visual results are rendered from Flux-GS with various Spherical Harmonics configurations. 
    }
    \label{fig:sh}
  
\end{figure}

\noindent\textbf{Analysis of Spherical Harmonic Decomposition:} 
Fig.~\ref{fig:sh} illustrates the decomposition of our Spherical Harmonic (SH) components, validating the efficacy of our enhanced first-order SH representation. We can find that the 0th-order SH in Flux-GS primarily reconstructs the base diffuse color and global illumination of the scene, while the 1st-order and $\Delta$1st-order SH components are specialized to capture high-frequency structural variations and refine local context. 
These results demonstrate that our proposed first-order SH strategy effectively captures intricate textures and complex specular, enabling high-quality rendering that closely approximates the ground truth,  even under mobile hardware.

\section{Conclusion}
In this work, we propose Flux-GS, which facilitates real-time rendering on resource-constrained mobiles by addressing the memory overhead of high-order Spherical Harmonics (SH). We present a Monte Carlo Specular Energy Aggregator to compress radiance energy into a low-order subspace, supplemented by an Attribute-Conditioned SH Enhancement module that refines the base SH with zero inference overhead. Additionally, a Multi-view Alpha-based Densification and Pruning strategy eliminates redundant Gaussians and enhances multi-view consistency. Evaluations demonstrate our method achieves competitive fidelity and real-time rendering speed. 
Flux-GS offers a scalable solution for efficient 3D scene representation, advancing spatial computing on everyday hardware.

\section*{Acknowledgements}
This research is funded in part by ARC-Discovery grant (DP220100800 to XY), ARC-DECRA grant (DE230100477 to XY), NVIDIA academic grant program, and Google Research Scholar Program. We thank all anonymous reviewers and ACs for their constructive suggestions.

\appendix

\section{Preliminary}

\noindent\textbf{Gaussian Splatting:}
3D Gaussian Splatting (3DGS)~\cite{kerbl2023gaussiansplatting} is an explicit Gaussian-based rendering method that represents a scene using a collection of anisotropic 3D Gaussians, enabling high-quality novel view synthesis. 
Each 3D Gaussian \( \mathcal{G}_i \) contains 3D geometry and appearance parameters, as \( (\mathbf{\mu}_i, \Sigma_i, o_i,c_i) \). 
For geometry representation, \( \mathbf{\mu}_i \in \mathbb{R}^3 \) defines the 3D position of the 3D Gaussian in its world coordinates, while the covariance matrix \( \Sigma_i \in \mathbb{R}^{3 \times 3} \), which is always positive semi-definite and symmetric, characterizes the spatial extent and orientation of the anisotropic Gaussian ellipsoid. 
In terms of geometric modeling, \( \Sigma_i \) is decomposed into a scale component \( s_i \) and a rotation component \( r_i \), where \( s_i \) controls the size of the Gaussian and \( r_i \) specifies its orientation.
As for appearance, each Gaussian leverages a color $ c_i(\omega) $ and an opacity value \( o_i \in [0,1] \), where Gaussian color \( c_i \) is represented using spherical harmonic coefficients \(c_i \).

To render an image, all 3D Gaussians are projected onto the image space using a standard perspective camera model. 
After that, a depth-sorting procedure is then applied to ensure that $\mathcal{N} $ Gaussians are rendered in a near-to-far order. Following this ordering, the final color of a pixel is computed via alpha blending:
\begin{equation}
    \mathbf{C} = \sum_{i=1}^{\mathcal{N}}  c_i(\omega) \alpha_i T_i, 
    \quad T_i = \prod_{j=1}^{i-1} (1 - \alpha_j), 
    \quad \alpha_i = o_i \exp\left(-\frac{1}{2}\Delta x_i^T \Sigma_i^{-1} \Delta x_i\right),
\end{equation}
where \( T_i \) denotes the accumulated transmittance before the \( i \)-th Gaussian, 
\( \Delta x_i = x_i - \mu_i \) represents the positional offset between the projected pixel location and the Gaussian mean.
\( \mathcal{N} \) is the total number of Gaussians contributing to the pixel.
Overall, Gaussian Splatting provides a differentiable, efficient, and compact representation for complex scenes, making it well-suited for real-time rendering.

\noindent\textbf{Spherical Harmonics:}
Spherical harmonics provide an orthonormal basis for square-integrable functions defined on the unit sphere $\mathbb{S}^2$. Any sufficiently smooth directional function $c(\boldsymbol{\omega}) : \mathbb{S}^2 \rightarrow \mathbb{R}^3$ can be expanded as
\begin{equation}
c(\boldsymbol{\omega}) = \sum_{\ell=0}^L \sum_{m=-\ell}^{\ell} c_{\ell m} Y_{\ell m}(\boldsymbol{\omega}),
\end{equation}
where $Y_{\ell m}$ denotes the real spherical harmonics basis functions, and $c_{\ell m}$ are the corresponding SH coefficients.
In practice, the expansion is truncated to a finite maximum degree $L$, yielding $(L+1)^2$ coefficients. In modern neural rendering systems, low-order SH (e.g., $L \leq 2$) are often preferred due to their compactness and numerical stability, while higher-order SH are sometimes used during training to capture high-frequency view-dependent effects.

\section{Additional Implementation}
As shown in the main paper, we train our proposed Flux-GS with 30k iterations, following the vanilla 3DGS, without pretraining or distillation. In the initial 3k iterations, we train with the third-order Spherical Harmonics. Then, we employ our proposed Monte Carlo Specular Energy Aggregator to obtain the first-order SH representation from the previous third-order SH.

To our proposed Monte Carlo Specular Energy Aggregator, we sample $K=2048$ uniform points on a unit sphere. We only exert it once to obtain low-order SH representation.
To obtain the first-order SH, we employ two MLPs initialized with one hidden layer (64 neurons) for the mapping.
As to our proposed Attribute-Conditioned SH Enhancement module, this module is parameterized by 4 layers of MLPs with ReLU activation. The hidden neurons for these MLPs are (128, 64, 32, 12), respectively. 

To our proposed Multi-view Alpha-based Densification and Pruning strategy.
We sample 6 multiple cameras for this strategy to make sure multi-view consistency.
For the loss thresholds, we set $\tau^+=0.1$ and $\tau^-=0.01$ to identify regions of well-reconstructed and poorly-reconstructed. 
As for the quantile for the importance and pruning, we set $Q_\tau^+=0.6$ and  $Q_\tau^-=0.1$.
For the quantization, we leverage the same method as the previous Mobile-GS~\cite{mobilegs}.
We train our proposed Flux-GS on the RTX 4090 GPU.
To foster reproducibility, we would release our code.

\section{Additional Experiments}

\subsection{Monte Carlo $K$ Sampling Point Analysis}

\begin{figure}[t!]
    \centering
    \includegraphics[width=0.7\linewidth]{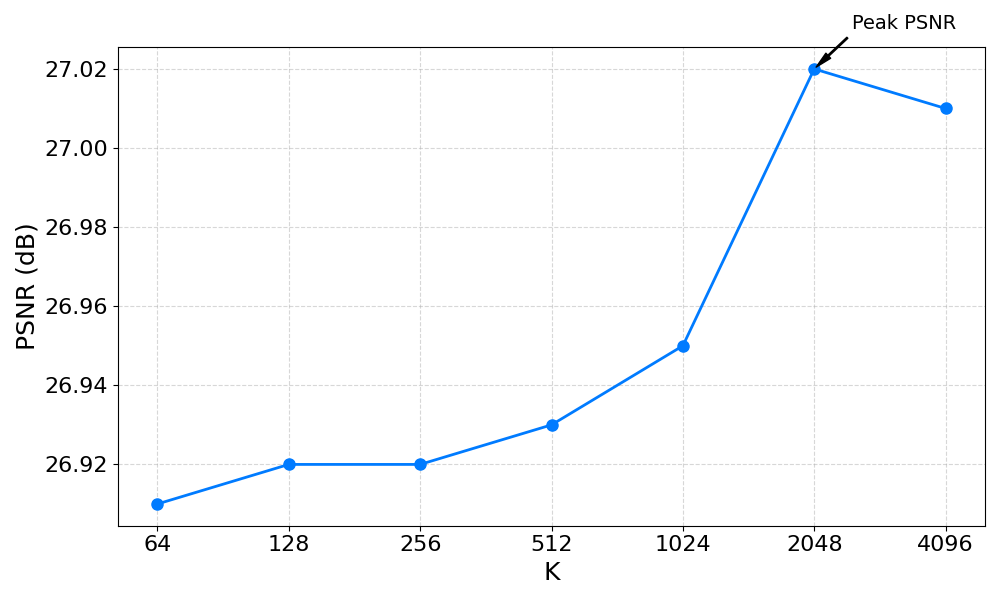}
    \caption{ \textbf{Impact of Monte Carlo Sampling Points $K$ on Reconstruction Quality.} We evaluate PSNR across varying sampling densities on the Mip-NeRF 360 dataset. Performance follows an upward trend as $K$ increases, with a notable saturation point appearing at $K = 2048$, indicating an optimal reconstruction fidelity.   }
    \label{fig:K}
\end{figure}

For sampling Density $K$, we investigate the sensitivity of our proposed method to the number of Monte Carlo sampling points $K \in \{64, \dots, 4096\}$. As illustrated in Fig.~\ref{fig:K}, the reconstruction quality PSNR demonstrates a monotonic improvement as $K$ increases from 64 to 2048. 
Specifically, we observe that increasing the density significantly reduces the variance inherent in the Monte Carlo estimator.
In particular, lower sampling rates ($K < 2048$) result in sub-optimal approximations of the integral.
Beyond this point ($K = 4096$), the gain in PSNR plateaus, suggesting that the estimator has converged.
Given the linear increase in computational overhead associated with larger $K$, we adopt $K =2048$ as the default configuration in our subsequent experiments to achieve a favorable rendering quality.

\subsection{Camera Count in Multi-view Alpha-based Densification}

\begin{figure}[t!]
    \centering
    \includegraphics[width=0.7\linewidth]{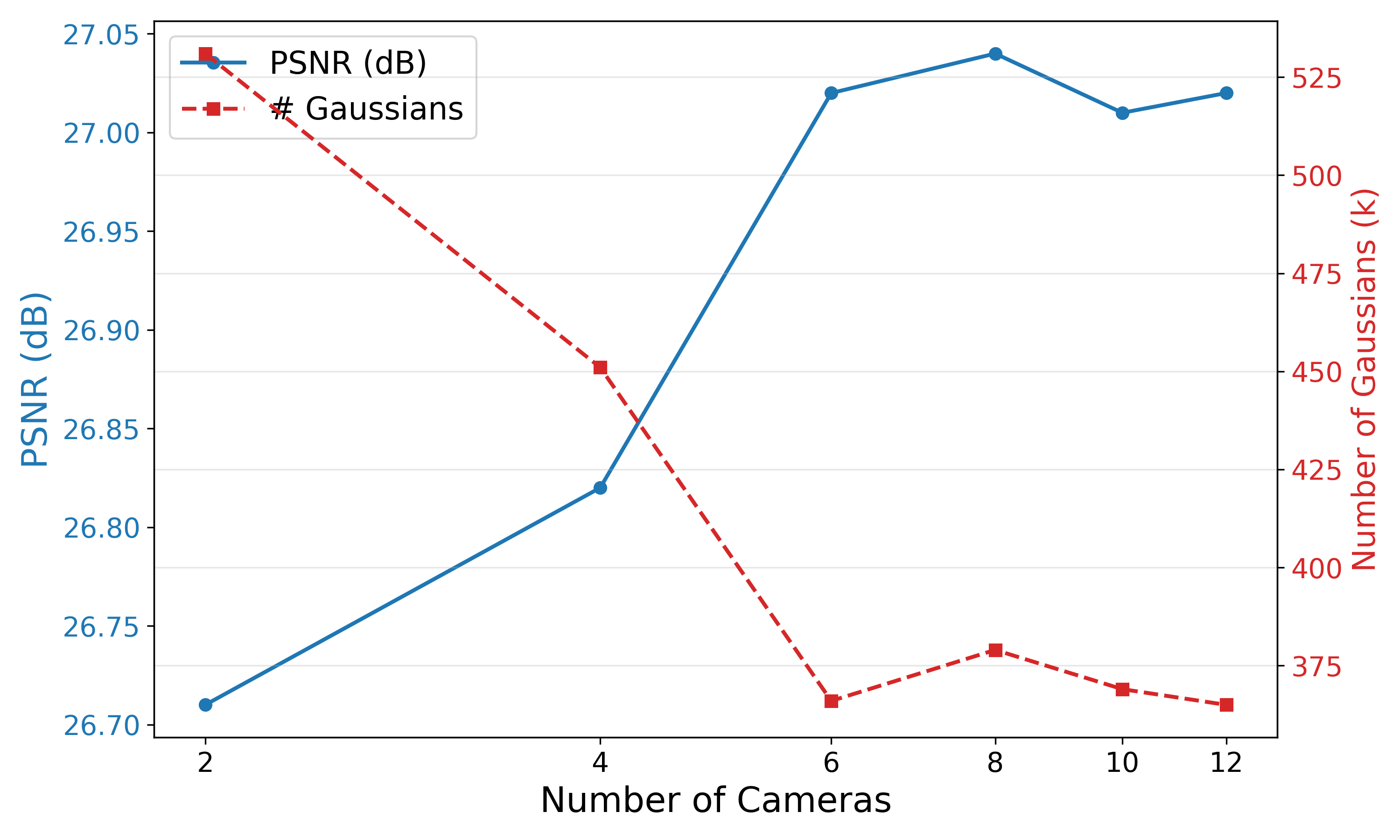}
    \caption{\textbf{Impact of the number of cameras in multi-view alpha-based densification.} We analyze the scaling impact of  varying camera counts in our multi-view densification strategy on the Mip-NeRF 360 dataset. While PSNR steadily improves and stabilizes with additional views, the total number of Gaussian primitives decreases significantly, demonstrating the effectiveness of our densification in avoiding aggressive densification while maintaining high-fidelity reconstruction.}
    \label{fig:camera_count}
\end{figure}

 To evaluate the robustness and efficiency of our multi-view densification module, we conduct an ablation study by varying the number of input camera views from 2 to 12. As illustrated in Fig.~\ref{fig:camera_count}, we observe a synergistic relationship between viewpoint coverage and representation efficiency. Specifically, increasing the camera count from 2 to 6 yields a significant PSNR improvement.
Crucially, as the number of views increases, the total Gaussian count drops significantly. This suggests that stronger multi-view constraints allow our densification strategy to more accurately localize geometric primitives for densification, effectively avoiding overfitting and redundancy. 
Therefore, we choose to sample 6 views for our multi-view densification, resulting in a superior trade-off between rendering performance and training burden.

\begin{table}[t!]
\caption{\textbf{Analysis of densification quantile $Q_\tau^+$ on the Mip-NeRF 360 dataset. }
Larger quantile hinders the densification of Gaussians, leading to fewer primitives and worse performance. 
}
\centering
\scriptsize
\def\arraystretch{1.2}
\begin{tabular}{l|ccccll}
\toprule
Quantile $Q_\tau^+$ & 0.4 & 0.5     & \textbf{0.6}    & 0.7    & 0.8    &0.9    \\  \midrule
PSNR $\uparrow$    & 27.14 & 27.11& \textbf{27.02}& 26.95& 26.94&26.91\\
Gaussian Points $\downarrow$ & 391667 & 381437& \textbf{366327}& 353215& 346141&331313\\ \bottomrule
\end{tabular}
\label{table:densify}
\end{table}
\begin{table}[t!]
\caption{\textbf{Analysis of Pruning quantile $Q_\tau^-$ on the Mip-NeRF 360 dataset. }
Larger quantile reduces the effect of pruning, leading to more primitives and better PSNR results. 
}
\centering
\scriptsize
\def\arraystretch{1.2}
\begin{tabular}{l|cccccc}
\toprule
Quantile $Q_\tau^-$ & 0.01& \textbf{0.1}& 0.3& 0.5& 0.7&0.9\\ \midrule
PSNR  $\uparrow$   & 26.81& \textbf{27.02}& 27.01& 27.04& 27.03&27.04\\
Gaussian Points $\downarrow$& 332315& \textbf{366327}& 373631& 377123& 382352&392521\\ \bottomrule
\end{tabular}
\label{table:prune}
\end{table}

\subsection{Densification and Pruning Quantile Analysis }
In our proposed multi-view alpha-based densification and pruning strategy, we utilize quantile $Q_\tau^+$ and $Q_\tau^-$ to identify Gaussians for densification and pruning, respectively. As reported in Table~\ref{table:densify} and \ref{table:prune}, we analyze the effect of these hyperparameters on rendering performance. It is obvious that when $Q_\tau^+$ becomes larger, fewer Gaussians are split and cloned, leading to worse results. To achieve a favorable trade-off between rendering performance and the number of Gaussian primitives, we set $Q_\tau^+ = 0.6$.
For the pruning threshold $Q_\tau^-$, a smaller value results in more aggressive pruning, removing a larger number of Gaussians and consequently harming rendering performance. Based on our analysis, we select $Q_\tau^- = 0.1$, which provides a better trade-off between model compactness and rendering quality.

\subsection{Per-Scene Results}
As shown in Table~\ref{tab:t_3drec_supp_psnr}, \ref{tab:t_3drec_supp_ssim}, and \ref{tab:t_3drec_supp_lpips}, we present per-scene evaluation results on the Mip-NeRF 360 dataset. The results demonstrate that our proposed Flux-GS can achieve comparable performance with 3DGS and other state-of-the-art light-weight GS variants. 
We further display training time comparisons in Table \ref{tab:t_3drec_supp_train}.
It is obvious that our proposed Flux-GS uses much less training time to obtain competitive rendering performance. These results further demonstrate the practicality of our method for real-time mobile rendering. This efficiency is primarily attributed to our proposed first-order SH design and multi-view regulated Gaussian variation, which significantly reduces the parameters of Gaussian primitives.

\begin{table*}[t!]
\centering 
\caption{\textbf{PSNR $\uparrow$ evaluation of state-of-the-art novel view synthesis methods for each scene on Mip-NeRF 360 dataset~\cite{mip-nerf360}}. The best results are highlighted.}
\scriptsize
\def\arraystretch{1.2}
\begin{adjustbox}{max width=\textwidth}
\begin{tabular}{lccccccccc}
\toprule
\textbf{Method}  & \textbf{bicycle} & \textbf{garden} & \textbf{stump} & \textbf{flowers} & \textbf{treehill} & \textbf{counter} & \textbf{kitchen} & \textbf{room} & \textbf{bonsai} \\
\midrule
3DGS~\cite{kerbl2023gaussiansplatting}  & \textbf{25.23} & \textbf{27.38} & 26.55 & 21.44 & 22.49 & 28.70 & 30.32 & 30.63 & \textbf{31.98} \\
Speedy-Splat~\cite{hanson2024speedy}  & 24.78 & 26.70 & 26.79 &  21.21  & 22.57  & 28.28 & 29.91 & 30.99 & 31.29 \\
Mobile-GS~\cite{mobilegs} & 24.91 & 26.65 & 26.82 &  21.41  &  \textbf{22.77} & \textbf{28.82} & \textbf{30.47} & 30.95 & 31.25 \\

 Flux-GS (\textbf{Ours})& 24.52& 26.66& \textbf{26.92}& \textbf{21.45}& 22.73& 28.53& 30.31& \textbf{31.21}&30.85\\  \bottomrule
\end{tabular}
\end{adjustbox}
\label{tab:t_3drec_supp_psnr}
\end{table*}

\begin{table*}[t!]
\centering \scriptsize \def\arraystretch{1.2}
\caption{\textbf{ SSIM $\uparrow$ evaluation of state-of-the-art novel view synthesis methods for each scene on Mip-NeRF 360 dataset~\cite{mip-nerf360}}. The best results are highlighted.}

\begin{adjustbox}{max width=\textwidth}
\begin{tabular}{lccccccccc}
\toprule
\textbf{Method} &  \textbf{bicycle} & \textbf{garden} & \textbf{stump} & \textbf{flowers} & \textbf{treehill} & \textbf{counter} & \textbf{kitchen} & \textbf{room} & \textbf{bonsai} \\
\midrule
3DGS~\cite{kerbl2023gaussiansplatting} & \textbf{0.765} & \textbf{0.864} & 0.770 & 0.602 & 0.633 & \textbf{0.907} & \textbf{0.925} & 0.918 & \textbf{0.940} \\
Speedy-Splat~\cite{hanson2024speedy} & 0.704 & 0.815 & 0.765 & 0.561  & 0.590   & 0.878 & 0.894 & 0.905 & 0.927 \\
Mobile-GS~\cite{mobilegs}& 0.740 & 0.823 & 0.777 & 0.593  &  \textbf{0.643}  & 0.905 & 0.920 & \textbf{0.924} & 0.936 \\

 Flux-GS (\textbf{Ours})& 0.751& 0.846& \textbf{0.778}& \textbf{0.605}& 0.641& 0.891& 0.911& 0.912&0.931\\  
\bottomrule
\end{tabular}
\end{adjustbox}
\label{tab:t_3drec_supp_ssim}
\end{table*}

\begin{table*}[t!]
\centering \scriptsize \def\arraystretch{1.2}
\caption{\textbf{LPIPS $\downarrow$ evaluation of state-of-the-art novel view synthesis methods for each scene on Mip-NeRF 360 dataset~\cite{mip-nerf360}}. The best results are highlighted.}
\begin{adjustbox}{max width=\textwidth}
\begin{tabular}{lccccccccc}
\toprule
\textbf{Method} &  \textbf{bicycle} & \textbf{garden} & \textbf{stump} & \textbf{flowers} & \textbf{treehill} & \textbf{counter} & \textbf{kitchen} & \textbf{room} & \textbf{bonsai} \\
\midrule
3DGS~\cite{kerbl2023gaussiansplatting}  & \textbf{0.211} & \textbf{0.108} & \textbf{0.217} & \textbf{0.339} & \textbf{0.329} & 0.201 & \textbf{0.127} & 0.220 & 0.205 \\
Speedy-Splat~\cite{hanson2024speedy}  & 0.333 & 0.213 & 0.288 &  0.419 & 0.463 & 0.260  &  0.198 & 0.260 & 0.231 \\
Mobile-GS~\cite{mobilegs} & 0.270 & 0.180  & 0.250 &  0.356  & 0.354 & \textbf{0.195} & 0.132 & 0.194 & 0.187 \\

 Flux-GS (\textbf{Ours})& 0.241& 0.163& 0.262& 0.361& 0.423& 0.221& 0.129& \textbf{0.192}&\textbf{0.176}\\   \bottomrule
\end{tabular}
\end{adjustbox}
\label{tab:t_3drec_supp_lpips}
\end{table*}



\begin{table*}[t!]
\centering \scriptsize \def\arraystretch{1.2}
\caption{\textbf{Training time (min)  $\downarrow$ evaluation of state-of-the-art novel view synthesis methods for each scene on Mip-NeRF 360 dataset~\cite{mip-nerf360}}. The best results are highlighted.}
\begin{adjustbox}{max width=\textwidth}
\begin{tabular}{lccccccccc}
\toprule
\textbf{Method} &  \textbf{bicycle} & \textbf{garden} & \textbf{stump} & \textbf{flowers} & \textbf{treehill} & \textbf{counter} & \textbf{kitchen} & \textbf{room} & \textbf{bonsai} \\
\midrule
3DGS~\cite{kerbl2023gaussiansplatting}  & 28& 31& 23& 21& 19& 24& 26& 23& 20\\
Speedy-Splat~\cite{hanson2024speedy}  & 17& 15& 14&  12& 13& 13&  15& 12& 14\\
Mobile-GS~\cite{mobilegs} & 106& 151& 143&  128& 146& 96& 74& 94& 84\\

 Flux-GS (\textbf{Ours})& \textbf{11}&\textbf{13}& \textbf{11}& \textbf{11}& \textbf{10}& \textbf{11}& \textbf{11}& \textbf{11}&\textbf{11}\\   \bottomrule
\end{tabular}
\end{adjustbox}
\label{tab:t_3drec_supp_train}
\end{table*}

\subsection{User Study}

\begin{figure}[t!]
    \centering
    \includegraphics[width=0.95\linewidth]{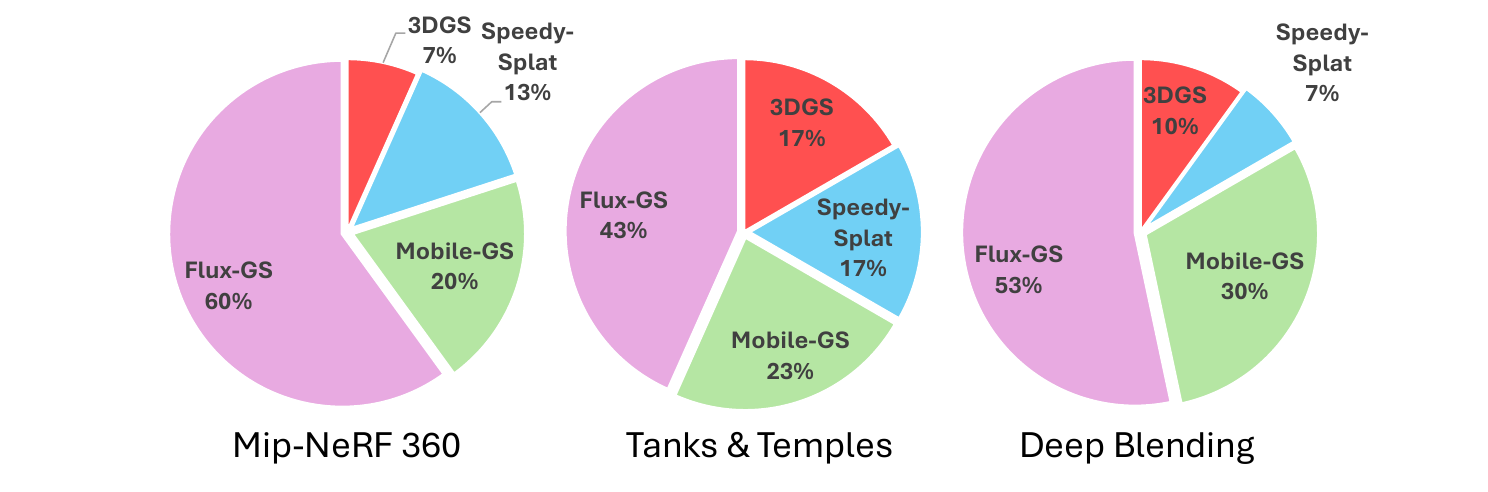}
    \caption{\textbf{User study of rendering quality.} We compare our Flux-GS with previous advanced methods, such as Mobile-GS~\cite{mobilegs}, Speedy-Splat~\cite{hanson2024speedy}, and 3DGS~\cite{kerbl2023gaussiansplatting}, in terms of subjective rendering quality.}
    \label{fig:user}
\end{figure}

We conducted a user study to evaluate the subjective rendering quality of Flux-GS, as illustrated in Fig.~\ref{fig:user}. We compared our method against Mobile-GS~\cite{mobilegs}, Speedy-Splat~\cite{hanson2024speedy}, and a quantized version of 3DGS \cite{kerbl2023gaussiansplatting}, ensuring a fair comparison across the Mip-NeRF 360 \cite{mip-nerf360}, Tanks \& Temples \cite{tanktemple}, and Deep Blending \cite{deepblending} datasets.
A total of thirty volunteers participated in the study and were asked to rate novel-view synthesis videos produced by each method.
The results indicate a clear preference for Flux-GS, which participants cited for its superior visual fidelity. This performance is largely due to our multi-view densification and pruning strategy, which effectively mitigate the floaters and rendering artifacts commonly seen in 3DGS. 
Overall, these results demonstrate that Flux-GS delivers high-quality and visually appealing renderings while remaining well-suited for resource-constrained mobile environments.

\section{Discussion and Limitations}
\subsection{Discussion}
Our proposed Flux-GS successfully bridges the gap between high-fidelity 3D scene representation and the strict hardware constraints of edge devices. While traditional 3D Gaussian Splatting (3DGS) achieves real-time rendering on desktop GPUs, its reliance on third-order Spherical Harmonics (SH) incurs severe memory bandwidth and storage overhead, making it unsuitable for mobile platforms. Flux-GS overcomes this by compressing appearance representation into the first-order SH.

The core success of our method lies in our proposed Monte Carlo Specular Energy Aggregator, which mathematically extracts the first-order directional moments of high-frequency specular residuals rather than directly discarding them. By projecting these residuals into a compact latent space with captured energy magnitude and direction, we avoid the expensive distillation processes or heavy pre-training typically required by previous lightweight methods. Furthermore, our Attribute-Conditioned SH Enhancement module acts as a lightweight geometry-conditioned corrector. Crucially, because these offsets are statically baked into the explicit Gaussian parameters prior to inference, this module introduces zero additional computational latency during the rasterization phase, securing high frame rates (FPS) on devices with the Snapdragon 8 Gen 3 GPU.

Finally, our Multi-view Alpha-based Densification and Pruning strategy improves the multi-view consistency and mitigates overfitting problems inherent to the standard single-view gradient-based densification. By leveraging stratified camera sampling, it imposes multi-view consistency to explicitly minimize the primitive budget without sacrificing structural integrity, representing a highly scalable solution for mobile and WebGL-based cross-platform rendering.

\subsection{Limitations}
While the Monte Carlo Specular Energy Aggregator successfully preserves visually salient lighting features, reducing the model to first-order SH inevitably loses some capacity to model highly complex, mirror-like, specular reflections that third-order SH can naturally capture.
Moreover, Flux-GS still optimizes full third-order SH for the initial 3,000 iterations before transitioning to the lower-order representation. 
Therefore, while the final model is remarkably lightweight, the initial training phase still requires peak memory consumption comparable to standard 3DGS.
The multi-view alpha-based pruning strategy relies on stratified camera sampling to evaluate primitive redundancy. If a scene contains extremely occluded or highly view-dependent micro-structures only visible from a narrow, unsampled angle, the multi-view guidance might incorrectly classify those important Gaussians as redundant and prune them.

\subsection{Future Works}
While Flux-GS significantly reduces the primitive count and SH footprint, introducing multi-view guidance in codebook-based compression or entropy coding could further push the storage limits down, enabling instantaneous streaming over low-bandwidth networks.
Extending Flux-GS to support dynamic scenes (4D Gaussian Splatting) on mobile platforms is deserved. Adapting the Multi-view Densification strategy to handle temporal redundancies could pave the way for real-time mobile playback of volumetric video.

%
%
\bibliographystyle{splncs04}
\bibliography{main}
\end{document}